%% file: main_arxiv2.tex
\documentclass[10pt,twocolumn,letterpaper]{article}

\usepackage[dvipsnames]{xcolor}
\usepackage{iccv}
\usepackage{tikz}
\usepackage{pgfplots}
\usepackage{times}
\usepackage{epsfig}
\usepackage{graphicx}
\usepackage{flushend}
\usepackage{amsmath,amssymb,mathtools}
\usepackage[ruled,vlined,linesnumbered]{algorithm2e}

\graphicspath{ {./images/} }

\newcommand{\vfunction}{\mathbf}

\def\R{\text{$\mathbb R$}}
\def\N{\text{$\mathcal N$}}
\def\Seeds{\Omega_\text{seeds}}
\def\dA{\text{$d_\text{\tiny A}$} }
\def\dB{\text{$d_\text{\tiny B}$} }
\def\fo{\text{$s$} }            %
\def\ff{\text{$\vfunction f$}}    %
\def\J{\text{$\pmb J$}}

\usepackage[pagebackref=true,breaklinks=true,letterpaper=true,colorlinks,bookmarks=false]{hyperref}

\iccvfinalcopy %

\def\iccvPaperID{5813} %

\ificcvfinal\fi

\newcommand{\NEW}{\color{red}}
\ificcvfinal\renewcommand{\NEW}{}\fi

\newcommand{\KL}{\operatorname{KL}}

\begin{document}

\title{Robust Trust Region for Weakly Supervised Segmentation}

\author{Dmitrii Marin\\
University of Waterloo, Canada\\
{\tt\small dmitrii.a.marin@gmail.com}
\and
Yuri Boykov\\
University of Waterloo, Canada\\
{\tt\small yboykov@uwaterloo.ca}
}

\maketitle
\ificcvfinal\thispagestyle{empty}\fi

\begin{abstract}

  Acquisition of training data for the standard semantic segmentation is expensive if requiring that each pixel is labeled. Yet, current methods significantly deteriorate in weakly supervised settings, \eg where a fraction of pixels is labeled or when only image-level tags are available. It has been shown that regularized losses---originally developed for unsupervised low-level segmentation and representing geometric priors on pixel labels---can considerably improve the quality of weakly supervised training. However, many common priors require optimization stronger than gradient descent. Thus, such regularizers have limited applicability in deep learning. We propose a new robust trust region approach\footnote{\scriptsize\url{https://github.com/dmitrii-marin/robust_trust_region}} for regularized losses improving the state-of-the-art results. Our approach can be seen as a higher-order generalization of the classic chain rule. It allows neural network optimization to use strong low-level solvers for the corresponding regularizers, including discrete ones.%
\end{abstract}

\section{Introduction}

Our paper proposes a higher-order optimization technique for neural network training. While focused on semantic 
image segmentation, our main algorithmic idea is simple and general - integrate the standard {\em trust region} principle 
into the context of {\em backpropagation}, \ie the chain rule. We reinterpret the classic chain rule: 
instead of the chain of gradients/derivatives for a composition of functions, we formulate the corresponding chain of hidden
optimization sub-problems. Then, inspired by the {\em trust region} principle, we can substitute  a standard 
linear approximation solver (gradient descent) at any chain with a better higher-order solver. In short,
we replace the classic differentiation chain rule by the trust region chain rule in the context of backpropagation.  

Our work is motivated by the well-known challenges presented to the gradient descent
by typical regularization losses or geometric priors/energies ubiquitous in the context of weakly-supervised 
or unsupervised segmentation. 
To validate our approach, we present semantic segmentation results improving the state-of-the-art in the challenging 
setting where the training data has only a fraction of pixels labeled.
The generality of our main principle (trust region chain rule) and our promising results for a difficult problem
encourage further research. In fact, this work applies trust region principle only to the last ``chain'' in the network. 
We discuss several promising extensions for future work.

The rest of the introduction is organized as follows. To create a specific context for our general approach to
network training, we review loss functions relevant for weakly-supervised or unsupervised segmentation. 
First, Sec.\ \ref{sec:reg_energy} discusses several standard geometric priors, regularization energies, 
clustering criteria, and their powerful solvers originally developed for low-level segmentation or general machine learning. 
Then, Sec.\ \ref{sec:reg_loss} outlines the use of such regularization objectives 
as losses for network training in the context of weakly supervised semantic (high-level) segmentation.
We also review the standard {\em trust region} principle (Sec.\ \ref{sec:tr_review}) and highlight our main 
contributions (Sec.\ \ref{sec:contributions}) based on the general idea of applying trust region 
(with powerful solvers) to network training.

\subsection{Regularized energies in low-level segmentation} \label{sec:reg_energy}
Assuming discrete segmentation $s\in \{1,2,\dots,K\}^N$ where $K$ is the number of categories and $N$ is the number of image pixels, one common low-level segmentation energy can be represented as  
\begin{equation} \label{eq:Potts}
    E(s) \; =\; -\sum_i \log P(I_i|s_i)\; + \sum_{\mathclap{\{i,j\}\in \N}} w_{ij}\,[s_i\neq s_j]
\end{equation}
where $I_i$ is a low-level feature (\eg intensity, color, texture) at pixel $i$ with distribution functions $P(\cdot|k)$ 
for each category $k$, neighborhood system $\N$ describes any pairwise connectivity (typically 4-, 8-grid \cite{BJ:01} or denser \cite{koltun:NIPS11}), weights $w_{ij}$ represent given pairwise affinities 
(typically Gaussian kernel for low-level features $I_i$ and $I_j$ \cite{BVZ:PAMI01,BJ:01,GrabCuts:SIGGRAPH04,koltun:NIPS11}), and
$[\cdot] $ is the Iverson bracket operator returning $1$ if the argument is true and $0$ otherwise.
The energy above combines the log-likelihoods term enforcing consistency with given (low-level) feature distributions
and a pairwise regularizer (Potts model) term enforcing geometric prior on shape smoothness with alignment to image intensity edges.

The Potts model has several efficient combinatorial \cite{BVZ:PAMI01} and LP-relaxation solvers \cite{Kolmogorov:PAMI06,kumar2009analysis}. Besides, 
there are many regularization objectives that are closely related to the first-order shape regularization 
in \eqref{eq:Potts}, but derived from a different discrete or continuous formulation of the low-level segmentation 
and equipped with their own efficient solvers, \eg geodesic active contours \cite{caselles1997geodesic}, snakes \cite{Kass:88}, 
power watersheds \cite{couprie2010power}, to name a few. Moreover, there are many other regularization terms going beyond 
the basic first-order smoothness (boundary length) enforced by the Potts term in \eqref{eq:Potts}. 
The extensions include curvature \cite{shekhovtsov2012curvature,part_enum:iccv13,Nieuwenhuis_2014_CVPR}, Pn-Potts \cite{kohli2009robust}, convexity \cite{convexity:pami17,kconvex:eccv18}, etc.

Common continuous formulations of the low-level segmentation use {\em relaxed} variable 
$s \in \Delta_K^N$ combining pixel-specific distributions $s_i = (s_i^1,\dots,s_i^K)\in\Delta_K$ over $K$ categories, where $\Delta_K$ is 
the {\em probability simplex}. 
In this case the segmentation objective/energy should also be relaxed, \ie, defined over real-values arguments.
For example, one basic relaxation of the Potts segmentation energy in \eqref{eq:Potts} is
\begin{equation} \label{eq:quadratic_Potts}
  -\sum_i \sum_k s_i^k \log P(I_i|k)\; + \sum_{\mathclap{\{i,j\}\in \N}} w_{ij}\,\|s_i-s_j\|^2
\end{equation}
using a linear relaxation of the likelihood term and a quadratic relaxation of the Potts model. Note that 
there could be infinitely many alternative relaxations. Any specific choice affects the properties of the relaxed solution, 
as well as the design of the corresponding optimization algorithm. 
For example, simple {\em quadratic} relaxation in \eqref{eq:quadratic_Potts} is convex suggesting simpler optimization,
but its known to be a non-tight relaxation of the Potts model \cite{ravikumar2006quadratic} 
leading to weaker regularization properties unrelated to geometry or shape. There are many better alternatives, \eg 
using different norms \cite{couprie2010power} or other convex formulations 
\cite{ChanNikEsid:05,chambolle2011first,chambolle2012convex}. The {\em bilinear} relaxation of the Potts term below
\begin{equation} \label{eq:bilinear_Potts}
  -\sum_i \sum_k s_i^k \log P(I_i|k)\; + \sum_k (1-s^k)^\top \,W\, s^k
\end{equation}
is tight \cite{ravikumar2006quadratic}, but it is non-convex and, therefore, more difficult to optimize. 
In the formula above, vector $s^k :=(s_i^k)$ combines segmentation variables for soft-segment $k$, 
and $N\times N$ affinity matrix $W_{ij}=w_{ij}\,[\{i,j\}\in \N]$ represents the neighborhood system $\N$ and 
all pairwise (\eg Gaussian) affinities $w_{ij}$ between image pixels. Note that Potts regularization
is closely related to the {\em Normalized cut} objective 
$\sum_k \frac{(1-s^k)^\top \,W\, s^k}{1^\top\,W\, s^k}$ for unsupervised segmentation \cite{Shi2000}.

It is common to combine energies like \eqref{eq:Potts},\eqref{eq:quadratic_Potts},\eqref{eq:bilinear_Potts} 
with constraints based on user interactions (weak supervision). 
While there are different forms of such supervision, the most basic one is based on adding the seed loss 
\cite{BJ:01} defined over pixels in subset $\Seeds$ with user-specified category labels $y_i$.
Assuming $s_i\in\Delta_K$, it can be written as a partial {\em cross entropy} (PCE) for pixels $i\in\Seeds$
\begin{equation} \label{eq:seeds}
   E_\text{seeds}(s)\;=\;  -\sum_{i\in \Seeds} \log s_i^{y_i}
\end{equation}
and, when restricted to {\em one-hot} $s_i$ representing hard segmentation, 
it reduces to the hard constraints over seeds \cite{BJ:01}. That is, for 
integer-valued $s_i\in\{1,\dots,K\}$ the seed loss is equivalent to 
$\sum_{i\in \Seeds} \lambda \,[s_i=y_i]$ for infinitely large $\lambda$.

The log-likelihood loss, \eg the first term in \eqref{eq:Potts} or \eqref{eq:bilinear_Potts},
is common in low-level segmentation and its importance cannot be underestimated. 
In basic formulations, the distributions of (low-level) features $P(\cdot|k)$ can be 
assumed given for each category $k$. However, if such distributions are not known {\em a priori}, 
their representation $P(\cdot|\theta_k)$ can explicitly include unknown distribution parameters $\theta_k$ 
for each category $k$. Then, the overall loss $E(s,\theta)$ adds $\theta=\{\theta_k\}$ as an extra variable. 
Optimization of $E(s,\theta)$ over both $s$ and $\theta$ corresponds to joint estimation of 
segmentation and {\em maximum likelihood} (ML) estimation of distribution parameters, as in well-known
unsupervised low-level segmentation formulations by Zhu \& Yuille \cite{Zhu:96} and Chan \& Vese \cite{Chan-Vese-01b}.
Similar ideas are also used in box-interaction methods~\cite{GrabCuts:SIGGRAPH04}.

\subsection{Regularized losses in DNN segmentation} \label{sec:reg_loss}

Unlike low-level segmentation methods based on readily available low-dimensional features 
(like color, texture, contrast edges), deep neural network (DNN) approaches to segmentation 
learn complex high-dimensional ``deep'' features that can discriminate semantic categories. 
Thus, one can refer to such methods as {\em high-level} segmentation, and to such learned features 
as {\em high-level} features. 

The most standard way to train segmentation networks is based on {\em full supervision} requiring 
a large collection of images where all pixels are accurately labeled. Such training data is expensive to get.
The training is based on minimizing the cross-entropy (CE) loss similar to the seed loss in low-level segmentation.
For simplicity focusing on a single training image, CE loss is
\begin{equation}\label{eq:ce_loss}
    E_{\text{\tiny CE}}(s(\theta)) \;=\;-\sum_i  \log s_i^{y_i} (\theta)
\end{equation}
where $s(\theta) = \ff (\theta) \in \Delta_K^N$ is the (relaxed) segmentation output of the network $\ff(\theta)$ 
with parameters $\theta$. For brevity, here and later in this paper we omit the actual test image 
from the arguments of the network function $\ff$.

The fundamental difference with low-level segmentation reviewed above is that instead of minimizing losses $E$ directly over 
segmentation variable $s$, now the optimization arguments are parameters $\theta$ of the network producing such segmentation. 
Estimating parameters $\theta$ can be interpreted as learning deep features. Note that this task is much more complex
than ML estimation of distribution parameters for $P(I|\theta)$ in low-level segmentation with fixed low-level features $I$, as reviewed above. This explains why network optimization requires a large set of fully labeled training images, 
rather then a single image (unlabeled or partially-labeled), as in low-level segmentation.

The goal of weakly supervised segmentation is to train the network with as little supervision as possible. First of all,
it is possible to train using only a subset of labeled pixels (seeds) in each image \cite{kolesnikov2016seed,NCloss:CVPR18} 
in exact analogy with \eqref{eq:seeds} 
\begin{equation}\label{eq:pce loss}
    E_{\text{\tiny PCE}}(s(\theta)) \;=\;-\sum_{i\in\Seeds}  \log s_i^{y_i} (\theta)
\end{equation}
In particular, as shown in \cite{NCloss:CVPR18}, this simple, but principled approach can outperform more complex 
heuristic-based techniques. To improve weakly-supervised training, it is also possible to use
standard low-level regularizes, as in Sec.\ \ref{sec:reg_energy}, that leverage a large number of unlabeled pixels  \cite{zheng2015conditional,kolesnikov2016seed,NCloss:CVPR18,Rloss:ECCV18,marin2019beyond}. For example,
\cite{Rloss:ECCV18} achieves the state-of-the-art 
using bilinear relaxation of the Potts model in \eqref{eq:bilinear_Potts}
\begin{equation}\label{eq:crf loss}
    E_{\text{Potts}}^{\text{bl}}  (s(\theta)) \;=\;\sum_k (1-s^k(\theta))^\top \,W\, s^k(\theta)
\end{equation}
as an additional regularization loss over all (including unlabeled) pixels. 
For some $\nu>0$, their continuous total loss
\begin{equation}\label{eq:total loss}
    E \;\;=\;\; E_{\text{\tiny PCE}} \; + \; \nu \, E_{\text{Potts}}^{\text{bl}}.
\end{equation}

More generally, standard regularization losses from low-level segmentation are commonly used 
in the context of segmentation networks. Such losses and their solvers are ubiquitous 
in weak-supervision techniques using seeds or boxes to generate fully-labeled {\em proposals} \cite{khoreva2017simple, scribblesup}. 
Optimization of low-level regularizers is also common for network's output post-processing, 
typically improving performance during testing \cite{deeplab}. Also,
the corresponding low-level solvers can be directly integrated as solution-improving 
layers \cite{zheng2015conditional}.

\subsection{Weakly supervised semantic segmentation}

Weak supervision for deep neural network semantic segmentation comes in many different forms, \eg image-level tags \cite{pathak2015constrained,papandreou2015weakly,kolesnikov2016seed}, scribbles/clicks \cite{scribblesup,NCloss:CVPR18,Rloss:ECCV18,marin2019beyond}, and bounding boxes \cite{papandreou2015weakly,khoreva2017simple,pmlr-v121-kervadec20a}. These works employ a large variety of strategies to compensate for the lack of labels. The concept of \textit{multiple instance learning} (MIL) naturally fits the weakly supervised setting. 
Since generic MIL methods produce small unsatisfactory segments, more specialized methods are needed. For example, methods \cite{pathak2015constrained,pmlr-v121-kervadec20a} impose constraints on the output of the neural network during learning. There are several segmentation-specific constraints, such as size bias, constraints on present labels, tightness \cite{lempitsky2009image}, \etc. \cite{kolesnikov2016seed,Rloss:ECCV18,marin2019beyond} incorporate edge alignment constraints. Proposal generation methods \cite{khoreva2017simple, scribblesup} aim to generate/complete the ground truth to use fully-supervised learning. However, DNNs are vulnerable to errors in proposals. More robust approaches use EM \cite{papandreou2015weakly} or ADMM~\cite{marin2019beyond} to iteratively correct errors in ``proposals''. 

Some related prior work on weakly supervised DNN segmentation \cite{scribblesup} uses some specific non-robust version 
of the joint loss related to our approach. Similar losses (studied in segmentation since 1980s) do not imply similar algorithms. In particular, they iterate explicit low-level segmentation of 
super-pixels \cite{felzenszwalb2004efficient} and pixel-level network training, where at each iteration 
the network is trained from scratch\footnote{That is, resetting the network to the ImageNet pre-trained parameters.} 
and to convergence. They motivate such integration by improved results only.
They also argue that ``when network gradually learns semantic content, the high-level information can help with the graph-based scribble propagation", suggesting their main focus on improved ``proposals". As shown in~\cite{NCloss:CVPR18,Rloss:ECCV18}, their method is outperformed by using only the partial cross entropy on seeds \eqref{eq:pce loss}.

\subsection{Classic \textit{\textbf{\large trust region}} optimization} \label{sec:tr_review}

{\em Trust region} is a general approximate iterative local optimization method \cite{boyd:04} allowing to use
approximations \underline{with good solvers} when optimizing arbitrarily complex functions. 
To optimize $g(x)$, it solves sub-problem 
$\min_{\|x-x_t\|\leq \epsilon} \tilde{g}(x)$ 
where function $\tilde{g}\approx g$ is an approximation that can be ``trusted'' in some region $\|x-x_t\|\leq \epsilon$ 
around the current solution. If $\tilde{g}$ is a linear expansion of $g$, this reduced to the gradient descent.
More accurate higher-order approximations can be trusted over larger regions allowing larger steps. The sub-problem is often formulated as unconstrained Lagrangian optimization
$\min_x \tilde{g}(x) + \lambda \|x-x_t\|$ where $\lambda$ indirectly controls the step size.

\subsection{Related optimization work and contributions} \label{sec:contributions}

The first-order methods based on stochastic gradient descent dominate deep learning due to their simplicity, efficiency, and scalability. However, they often struggle to escape challenging features of the loss profile, \eg ``valleys'', as the gradients lack information on the curvature of the loss surface.  Adam \cite{kingma2014adam} combines gradients from many iterations to gather such curvature information. On the other hand, the second-order methods compute parameters update in the form $\Delta\theta = H^{-1} \nabla_\theta E(\ff(\theta))$, \cf \eqref{eq:class_cr}, where $H$ is the Hessian or its approximation. In neural networks, computing the Hessian is infeasible, so various approximations are used, \eg diagonal or low-rank \cite{Bishop06}. The efficient computation of Hessian-vector products is possible \cite{pearlmutter1994fast, schraudolph2002fast}; while solving linear systems with Hessian is still challenging \cite{sutskever2013importance}. Another group of methods is based on employing Gaussian-Newton matrix and K-FAC approximations \cite{martens2015optimizing,ba2016distributed,pmlr-v70-botev17a,osawa2019large}.

{\NEW
Our approach is related to the proximal methods~\cite{martinet1970breve}, in particular to the \textit{proximal backpropagation}~\cite{frerix2018proximal} and \textit{penalty method}~\cite{carreira2014distributed}. In these works, the ``separation'' of the gradient update into implicit layer-wise optimization problems is formulated as a gradient update of a certain energy function. Taylor \etal~\cite{taylor2016training} use \textit{ADMM} splitting approach to separate optimization over different layers in distributed fashion. These works focus on neural network parameter optimization replacing backpropagation altogether. In contrast to \cite{carreira2014distributed,taylor2016training,frerix2018proximal}, we are primarily focused on optimization for complex loss functions in the context of the weakly supervised semantic segmentation, see Sec.\ref{sec:reg_loss}, while others focus on replacing the backpropagation in the intermediate layers. Also, unlike us, these methods use the squared Euclidean norm in their proximal formulations. Chen and Teboulle~\cite{chen1993convergence} generalize the proximal methods to \textit{Bregman divergences}, a more general class of functions which includes both the Euclidean distance and KL-divergence. Nesterov in~\cite{nesterov2020inexact} uses the Euclidean norm with a higher power improving the convergence of the proximal method.}

Our contribution are as follows:
\begin{itemize}
    \item New trust region optimization for DNN segmentation integrating higher-order low-level solvers into training. Differentiability of the loss is not required as long as there is a good solver, discrete or continuous. The classic differentiation chain rule is replaced by the trust region chain rule in the context of backpropagation.
    
    \item The local optimization in trust region framework allows to use arbitrary metrics, instead of Euclidean distance 
    implicit for the standard gradient descent. We discuss different metrics for the space of segmentations and 
    motivate a robust version of KL-divergence. 
    
    \item  We show benefits of our optimization for regularization losses in weakly supervised DNN segmentation, compared to the gradient descent. We set new state-of-the-art results for weakly supervised segmentation with scribbles achieving consistently the best performance at all levels of supervision, \ie from point-clicks to full-length scribbles. %
\end{itemize}

\section{Trust region for loss optimization} \label{sec:tr}

{\em Backpropagation} is the dominant method for optimizing network losses during training. It represents the gradient descent 
with respect to model parameters $\theta$ where the gradient's components are gradually accumulated using the classic 
{\em chain rule} while traversing the network layers starting from the output directly evaluated by the loss function. 

Motivated by the use of hard-to-optimize regularization losses (Sec.~\ref{sec:reg_energy}) in the context of weakly-supervised segmentation (Sec.~\ref{sec:reg_loss}), we propose higher-order {\em trust region} approach to network training. 
While this general optimization approach can be developed for any steps of the backpropagation (\ie chain rule) 
between internal layers, we focus on the very first step where the loss function is composed with the network output
\begin{equation} \label{eq:Ef}
\min_{\theta\in\R^m}  \;\;  E(\ff(\theta))  
\end{equation}
where some scalar loss function 
$$E: \R^n\rightarrow \R^1$$ is defined over $n$-dimensional output of a network/model $$\ff:\R^m \rightarrow \R^n.$$ 
Since during training the network's input is limited to fixed examples, 
for simplicity we restrict the arguments of network function $\ff$ to its training parameters $\theta\in \R^m$. 
Also note that, as a convention, this paper reserves the boldface font for vector functions (\eg network model $\ff$) 
and for matrix functions (\eg model's Jacobian $\J_\ff$). 

The main technical ideas of the trust region approach to network optimization \eqref{eq:Ef} in this section 
are fairly general. However, to be specific and without any loss of generality, this and (particularly) later sections 
can refer to the output of the network as {\em segmentation} so that
$$\R^n = \R^{N \times K} $$
where $N$ is the number of image pixels and $K$ is the number of distinct semantic classes. This is not essential.

Our general trust region approach to \eqref{eq:Ef} can be seen as a higher-order extension of the classic chain rule 
for the composition $E\circ \ff$ of the loss functions $E$ and model $\ff$. For the classic chain rule in the standard backpropagation 
procedure, it is critical that both $E$ and $\ff$ are differentiable. In this case, the classic chain rule for the 
objective in \eqref{eq:Ef} gives the following gradient descent update for parameters $\theta$ 
\begin{equation} \label{eq:class_cr}
\Delta\theta\;\;=\;\;-\alpha\; \nabla E^\top \,\J_\ff
\end{equation}
where $\Delta\theta\equiv\theta-\theta_t$ is an update of the model parameters from the current solution, $\alpha$ is the learning rate, $\nabla$ is the gradient operator, and $\J_\ff$ is the model's {\em Jacobian} $$\J_\ff:=\left[\frac{\partial \ff_i}{\partial \theta^j}\right].$$

We would like to rewrite the classic chain rule \eqref{eq:class_cr} in an equivalent form explicitly using 
a variable for segmentation $\fo\in\R^n$, which is an implicit (hidden) argument of the loss function $E$ in \eqref{eq:Ef}. 
Obviously, equation \eqref{eq:class_cr} is equivalent to two separate updates 
for the segmentation $\Delta\fo \equiv \fo-\fo_t$ and for the model parameters $\Delta\theta\equiv\theta-\theta_t$
\begin{align}
   \Delta \fo  \;\;\;= \;\; & \; - \alpha \;\, \nabla E^\top  \label{eq:cr:s1} \\[2ex]
   \Delta\theta\;\;\;=\;\; & \;\;\;\;\Delta \fo\;\,\J_\ff    \label{eq:cr:s2}
\end{align}
where the gradient $\nabla E$ is computed at the current segmentation $\fo_t := \ff(\theta_t)$.
Note that $\fo\in\R^n$ represents points (\eg segmentations) 
in the same space as the network output $\ff(\theta)\in\R^n$, the two should be clearly distinguished in the discourse. 
We will refer to $\fo$ as (explicit) segmentation {\em variable}, while $\ff(\theta)$ 
is referred to as segmentation {\em output}.

The updates in \eqref{eq:cr:s1} and \eqref{eq:cr:s2} correspond to two distinct optimization sub-problems. 
Clearly, \eqref{eq:cr:s1} is the gradient descent step for 
the loss $E(s)$ locally optimizing its linear Taylor approximation 
$\tilde{E}_{\text{linear}}(\fo)=E(\fo_t) + \nabla E^\top \Delta \fo$ 
over (explicit) segmentation variable $\fo\in B(\fo_t)\subset R^n$ 
in a neighborhood (ball) around $\fo_t$ 
\begin{equation} \label{eq:lowE}
 \fo_{t+1}  \;\;= \;\; \;\arg \min_{\mathclap{\fo\in B(\fo_t)}}\; \tilde{E}_{\text{linear}}(\fo) .
\end{equation}
While less obvious, it is easy to verify that $\theta$-update in \eqref{eq:cr:s2} is exactly 
the gradient descent step
\begin{equation}
\Delta {\theta} \; =\; -\frac12 \nabla_{\!\theta}\; \|\fo_{t+1}-\ff(\theta)\|^2  \label{eq:gradLS}
\end{equation}
corresponding to optimization of the least-squares objective
\begin{equation}
\min_{\theta} \; \|\fo_{t+1}-\ff(\theta)\|^2   \label{eq:LS}
\end{equation}
based on the solution $\fo_{t+1} \equiv \Delta\fo + \ff(\theta_t)$ for problem \eqref{eq:lowE}.

Our trust region approach to network training \eqref{eq:Ef} is motivated by the principled separation of 
the chain rule \eqref{eq:class_cr} into two sub-problems \eqref{eq:lowE} and \eqref{eq:LS}. 
Instead of the gradient descent, low-level optimization of the loss 
in \eqref{eq:lowE} can leverage powerful higher-order solvers available for many popular loss functions, 
see Sec.~\ref{sec:reg_energy}. In particular, the majority of common robust loss functions for unsupervised 
or weakly-supervised computer vision problems are well-known to be problematic for the gradient descent. 
For example, their robustness (boundedness) leads to {\em vanishing gradients} and sensitivity to local minima. 
At the same time, the gradient descent can be left responsible for the least-squares optimization in \eqref{eq:LS}. 
While it is still a hard problem due to size and non-convexity of the typical models $\ff(\theta)$, 
at least the extra difficulties introduced by complex losses $E$ can be removed into a different sub-problem.

Formally, our trust-region approach to training \eqref{eq:Ef} generalizes our interpretation of the classic chain rule 
in sub-problems \eqref{eq:lowE} and \eqref{eq:LS} as shown in \underline{iterative}  stages {\sc A}, {\sc B}: 
\begin{align}
\text{\sc stage A} \;\;\;\;   & \;\;\;\; (\text{\em low-level optimization}) \nonumber    \\[1ex]
   \fo_{t+1}  \;\;= \;\; & \;\arg \min_\fo \;\; \tilde{E}(\fo)\;+\;\lambda\;\dA(\fo,\ff(\theta_t)) \quad \label{eq:tr:s1} \\[2ex]
\text{\sc stage B} \;\;\;\;  & \;\; (\text{\em network parameters update}) \nonumber \\[1ex]
            & \;\;\;\;\;\;\;\;\, \underbrace{\min_\theta \;\; \dB(\fo_{t+1},\ff(\theta))}_{ \Downarrow } \label{eq:tr:s2} \\[-0.5ex]
    \Delta\theta  \;\; = \;\; & \;\;\;\;-\;\gamma \;\nabla_{\!\theta}\;  \dB(\fo_{t+1},\ff(\theta))  \label{eq:tr:grads2} 
\end{align}
where $\tilde{E}$ is some loss approximation, $\dA$ and $\dB$ are some distance/divergence measures.
Instead of $\alpha$ in \eqref{eq:cr:s1} and fixed weight $\frac{1}{2}$ in \eqref{eq:gradLS}, 
the overall learning speed of our training procedure is controlled by two parameters: 
(A) scalar $\lambda$ indirectly determining the step size 
from the current solution $\fo_t=\ff(\theta_t)$ in \eqref{eq:tr:s1}, and (B) scalar $\gamma$ defining the step size 
for the gradient descent in \eqref{eq:tr:grads2}. While both $\lambda$ and $\gamma$ are important for the learning speed, 
we mostly refer to $\lambda$ as a trust region parameter, while the term {\em learning rate} is reserved primarily 
for parameter $\gamma$ in \eqref{eq:tr:grads2}, as customary for the gradient descent step size in network optimization.
Note that similarly to the gradient descent \eqref{eq:class_cr}, stages A/B are iterated until convergence. While it is sensible 
to make several B-steps \eqref{eq:tr:grads2} in a row, in general, it is not necessary to wait for convergence 
in sub-problem \eqref{eq:tr:s2} before the next A-step.

Our formulation offers several significant generalizations of the classic chain rule.
{\bf First}, instead of the linear approximation \eqref{eq:lowE} implied by the {\em gradient descent} \eqref{eq:cr:s1}, 
we target higher-order approximations of the loss $\tilde{E}$ in \eqref{eq:tr:s1}. 
In some cases, the exact loss $E$ could be used\footnote{Note that parameter $\lambda$ in \eqref{eq:tr:s1}
controls two properties: the size of the trust region for approximation $\tilde{E}$, as well as the network's training speed.
While using exact loss $\tilde{E} = E$ implies that the trust region for such ``approximation'' should be the whole domain 
(\ie $\lambda=0$), the competing interest of limiting the training speed in \eqref{eq:tr:s2} may require $\lambda>0$.}\@.  
The corresponding powerful low-level solvers for \eqref{eq:tr:s1} are readily available for many types of 
useful robust losses, see Sec.~\ref{sec:reg_energy}. Note that for exact solvers when $\tilde{E}=E$, one may argue for
$\lambda=0$ allowing the network to learn from the best solutions for regularized loss $E$ implying global optima
in \eqref{eq:Ef}. However, such fixed proposals (Sec.~\ref{sec:reg_loss}) may result in overfitting to mistakes due
to well-known biases/weaknesses in common regularizers. Constraining loss optimization \eqref{eq:Ef} 
to the network output manifold in $\R^n$ motivates $\lambda>0$ in \eqref{eq:tr:s1}. 
More discussion is in Sec.\ \ref{sec:discussion}.

{\bf Second}, besides continuous/differentiable losses required by the standard backpropagation (chain rule), 
our trust region approach (stages A/B) allows training based on losses defined over discrete domains. 
There are several reasons why this extension is significant. For example, besides continuous solvers, optimization in 
\eqref{eq:tr:s1} now can use a significantly larger pool of solvers including many powerful discrete/combinatorial methods.
Moreover, this approach enables training of models with discrete decision functions, e.g. {\em step function} instead of 
{\em sigmoid}, or {\em hard-max} instead of the {\em soft-max}. This is further discussed in Sec.\ \ref{sec:discussion}.

{\bf Third}, the standard gradient descent \eqref{eq:class_cr} 
is implicitly defined over Euclidean metric, that manifests itself in our equations \eqref{eq:lowE} and \eqref{eq:LS} 
via the local neighborhood topology (Euclidean ball $B$) and the least-squares objective (squared Euclidean distance). 
In contrast, when replacing ball $B(\fo_t)$ in \eqref{eq:lowE} by the {\em trust region} term 
in \eqref{eq:tr:s1}, we explicitly define the trust region ``shape'' using function $\dA$. It could be any 
application-specific distance metric, quasi- or pseudo-metric, divergence, \etc. Similarly,
any appropriately motivated distance, distortion, or divergence function \dB in \eqref{eq:tr:s2} 
can replace the least squares objective in \eqref{eq:LS}. 

On the negative side, our trust region formulation could be more expensive due to the computational costs 
of the low-level solvers in stage A. 
In practice, it is possible to amortize stage A over multiple iterations of stage B.

\section{Robust metric for trust region}

The choice of metrics \dA and \dB defining the shape of the trust region above is application dependent. In the case of segmentation, the output of a neural network is typically obtained via the soft-max function. Hence, the space, in which the trust region operates, is the space of multiple categorical distributions over $K$ categories: $\Delta_K^N$.

Below, we generally discuss (robust) metrics over pairs of arbitrary probability distributions $p$, $q$ in  $\Delta_K^N$.
The goal of this section is to motivate our choice of metrics \dA and \dB in problems \eqref{eq:tr:s1}, \eqref{eq:tr:s2}
so that distribution $p$ can be associated with the segmentation variable $\fo$, 
and distribution $q$ can be associated with the network output $\ff(\theta)$. Besides this connection,
the following discussion of metrics over probability distributions is independent of the context of networks.

Note, metrics \dA or \dB do not have to be proper distances for the purposes of trust region optimization. Instead, one may use any divergence measure defined on space $\Delta_K^N$. Let us consider the Kullback–Leibler divergence:
\begin{align}
    \KL(p \| q) = & \sum_{i=1}^N \sum_{l=1}^K p^l_i \log \frac{p^l_i}{q^l_i} \notag %
     = -\sum_{i=1}^N \sum_{l=1}^K p^l_i \log q^l_i - H(p)
\end{align}
where $p, q \in \Delta_K^N$, and $p^l_i$ is the probability of pixel $i$ to have label $l$, and $H(p)$ is the entropy of distribution $p$.

A practically important case is when the distribution $p$ is degenerate or one-hot, \ie for each pixel $i$ there exists label $y_i$ such that $p^{y_i}_i = 1$ and for any label $k\neq y_i$ probability $p^k_i = 0$. In that case $H(p) = 0$ and
\begin{equation}\label{eq:kl-nll}
    \KL(p \| q) \;=\; \sum_i -\log q_i^{y_i},
\end{equation}
which is the cross-entropy or negative log-likelihood, a standard loss when $q$ is the probability estimate outputted by a neural network. In the following we assume \eqref{eq:kl-nll}.

\begin{figure}
    \centering
    \input{images/graphical_model}
    \caption{The unknown true labeling $Z$ corresponds to observed image $I$. The observed labeling $Y$ is assumed to be generated from the true $Z$ by a simple corruption model \eqref{eq:error model}.}
    \label{fig:error model}
\end{figure}
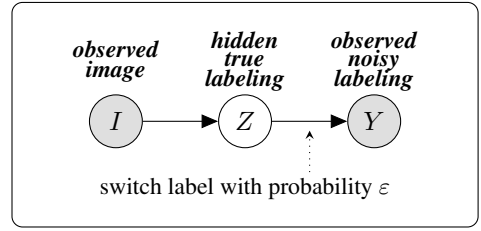

\begin{figure}
    \centering
    \input{images/robust_plot}
    \caption{{Robust loss as function of logits $x_i$}. There are $K=2$ classes; the ground truth label is $y_i=1$. If the current prediction $q^1_i$ is confident and does not coincide with $y_i$, see $x_i \ll 0$ or $q^1_i \approx 0$ on the plot, robust loss (b) becomes flatter avoiding the over-penalize in case of mistakes in the ground truth. In contrast, standard cross-entropy (a) behaves linearly, which may be detrimental to learning if the ground truth is mistaken.}
    \label{fig:robust plot}
\end{figure}
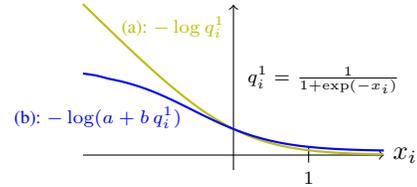

During the trust region procedure, intermediate solutions generated by a solver in \eqref{eq:tr:s1} may have a noticeable amount of misclassified pixels. It is known that many standard losses for neural networks, including cross-entropy \eqref{eq:kl-nll}, can result in training sensitive to idiosyncrasies in the datasets including mistakes in the ground truth \cite{geman1992neural,manwani2013noise,frenay2013classification}. Therefore, a robust distance measure may be needed. Our experiments show that robustness is critical. We propose a simple error model depicted in graphical model in Fig.~\ref{fig:error model}. Let random variable $Y_i$ be the observed noisy label of pixel $i$ and $Z_i$ be its hidden true label. We assume that the probability of observing label $l$ given true label $k$ is
\begin{equation}\label{eq:error model}
    \Pr(Y_i=l\,|\,Z_i = k) = \begin{cases}  
        1 - \varepsilon, & l = k,\\
        \frac{\varepsilon}{K - 1},& l\neq k, 
    \end{cases}
\end{equation}
where $\varepsilon$ is called the \textit{outlier probability} \cite{larsen1998design}.
The probability of pixel $i$ having label $l$ given image $I$ is
\begin{align}
 \Pr(&Y_i\!=\!l|I) %
     \;=\; \sum_{z=1}^K \Pr(Y_i\!=\!l | Z_i\!=\!z)\Pr(Z_i\!=\!z|I) \; = \notag\\
     &= a + b \, \Pr(Z_i\!=\!l|I)
\end{align}
where $a = \frac{\varepsilon}{K-1}$ and $b = 1 - K \, a$. The  probability $\Pr(Z_i\!=\!z|I)$ is unknown and is replaced by probability estimate $q_i^{l}$ yielding a robust version of divergence \eqref{eq:kl-nll}:
\begin{equation} \label{eq:robust loss}
    \sum_i -\log \left( a + b\,q_i^{y_i} \right).
\end{equation}
Figure \ref{fig:robust plot} compares cross-entropy \eqref{eq:kl-nll} with robust loss~\eqref{eq:robust loss}.

\begin{figure}
    \begin{center}
       \rotatebox{90}{~~~~~~~~~~~~~~~~Accuracy} \includegraphics[width=0.8\linewidth]{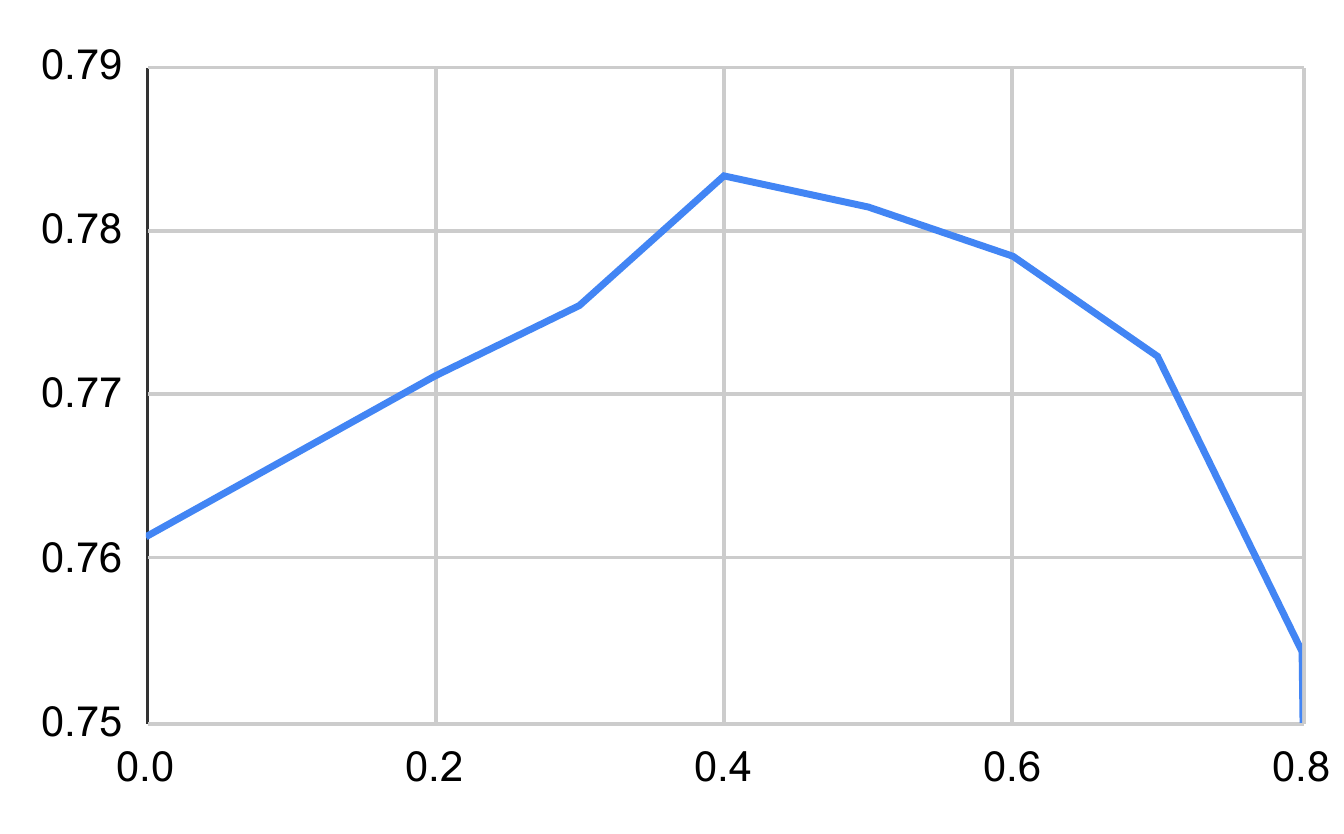}\\[-1ex]
       ~~~~~~robustness parameter $\varepsilon$
    \end{center}
    \caption{Classification accuracy on Fashion-MNIST dataset \cite{xiao2017/online} with noisy labels using a network with two convolutional, two fully-connected layers and \emph{robust loss} \eqref{eq:robust loss}. The original labels were uniformly corrupted with probability $\frac12$. The best accuracy is achieved at $\varepsilon=0.4$, which is close to the actual noise level.}
    \label{fig:fashion mnist}
\end{figure}

Our robust cross-entropy \eqref{eq:robust loss} is related to a more general approach for classification \cite{patrini2017making, sukhbaatar2015training}. In \cite{patrini2017making}, the corresponding robust cross-entropy (\emph{forward correction}) is
\begin{equation}\label{eq:forward procedure}
    \sum_i -\log \tilde q_i^{y_i} 
\end{equation}
where $\tilde q_i = T^\top q_i$, and $q_i$ is the vector of probability estimates at pixel $i$, and $T=[T_{lk}]$ is the \textit{noise transition matrix}: $T_{lk} = \Pr(Y=k\,|\,Z = l)$. 
The effect of different $\varepsilon$ is shown in example in Fig.~\ref{fig:fashion mnist}.

In practice, different pixels require different values of $\varepsilon$ in \eqref{eq:error model}. For example, in the scribble-based weakly supervised segmentation, the labels of seed pixels $\Seeds$ are known for sure. So, $\varepsilon=0$ for such pixels, and $\varepsilon>0$ for all other pixels. Thus, the robust ``metric'' is
\begin{equation} \label{eq:kl mixed}
    \KL_{\varepsilon,\Seeds}(p \| q) = \sum_{\mathclap{i \not\in \Seeds}} -\log \left( a + b\,q_i^{y_i} \right) + \sum_{\mathclap{i \in \Seeds}} -\log  q_i^{y_i}.
\end{equation}

In sum, we propose the following robust metrics for the trust region iterations \eqref{eq:tr:s1} and \eqref{eq:tr:grads2}:
\begin{equation}\label{eq:robust tr metrics}
    \begin{cases}
    \dA(p,q) \;=\; \KL(p \| q),  \\
    \dB(p,q) \;=\; \KL_{\varepsilon,\Seeds}(p \| q).
    \end{cases}
\end{equation}

\section{Results in weakly supervised segmentation}

To validate our approach (\ref{eq:tr:s1}-\ref{eq:tr:grads2}) we use standard efficient discrete solvers \cite{BVZ:PAMI01} 
for the loss 
\begin{equation}\label{eq:grid-tr loss}
    \tilde{E} \;\;=\;\; E_{\text{\tiny PCE}} \;+\; E_{\text{\tiny Potts}}
\end{equation}
where $E_{\text{\tiny Potts}}(s)=\sum_{{\{i,j\}\in \N}} w_{ij}\,[s_i\neq s_j]$ 
is the second (regularization) term in standard low-level energy \eqref{eq:Potts}. In this case, optimization in \eqref{eq:tr:s1} is limited 
to the corners of the simplex where $E_{\text{\tiny PCE}}$ reduces to the hard constraints 
over the seeds. In (\ref{eq:tr:s1}-\ref{eq:tr:grads2}) we use robust metrics \eqref{eq:robust tr metrics}. The overall method is
summarized in Alg.\,\ref{alg:robust tr}.

One natural baseline for Alg.\,\ref{alg:robust tr} is a standard method based on stochastic gradient descent (SGD) for regularized loss \eqref{eq:total loss} 
proposed in \cite{Rloss:ECCV18}, see Sec.\,\ref{sec:reg_loss}. 
Indeed, $E_{\text{\tiny Potts}}^{\text{\tiny bl}}$ is a relaxation of $E_{\text{\tiny Potts}}$, as discussed in Sec.\,\ref{sec:reg_energy}. 
Thus, \eqref{eq:total loss} is a relaxation of \eqref{eq:grid-tr loss}. Alg.\,\ref{alg:robust tr} with combinatorial 
solver for $\tilde{E}$ in \eqref{eq:grid-tr loss} can be seen as a {\em discrete} trust region approximation for \eqref{eq:total loss}.
In general, our approach (\ref{eq:tr:s1}-\ref{eq:tr:grads2}) allows other discrete or continuous solvers and/or
other approximations $\tilde{E}$.

First, PCE-GD baseline is the standard SGD optimizing partial cross-entropy \eqref{eq:pce loss}. It has been shown in \cite{Rloss:ECCV18,NCloss:CVPR18} that such approach outperforms more complex proposal (fake ground truth) generation methods such as \cite{scribblesup}. 
Second, Grid-GD is the SGD over regularized loss \eqref{eq:total loss} where the CRF neighbourhood is the standard 8-grid. Third, Dense-GD is the approach of \cite{Rloss:ECCV18} that uses the common fully-connected (dense) Potts CRF of \cite{koltun:NIPS11}. 

We use the ScribbleSup \cite{scribblesup} annotations for Pascal VOC 2012 \cite{pascal-voc-2012} dataset. ScribbleSup supplies scribbles, \ie a small subset of image pixels ($\approx 3$\%) is labeled while the vast majority of pixels is left unlabeled.

\begin{algorithm}[b]
\SetAlgoLined
 Initialize model \ff\ using ImageNet pretraining \;
 Tune parameters $\theta$ of model \ff\ by optimizing PCE-GD loss \eqref{eq:pce loss} \;
 Initialize $\gamma$ with the base learning rate \;
 \Repeat{required number of epochs is reached}{
  \For{each image in dataset}{
     compute segmentation variable \fo\ via \eqref{eq:tr:s1} using metric \dA in \eqref{eq:robust tr metrics} and loss \eqref{eq:grid-tr loss}\;
  }
  \For{$M$ epochs}{
   \For{each image (batch) in dataset}{
     update the network parameters $\theta$ using stochastic gradient descent for loss~\eqref{eq:tr:s2} with robust metric \dB in \eqref{eq:robust tr metrics} \;
     update rate $\gamma$ in accord with schedule\;
    }
  }
 }
 \caption{Robust Trust Region for Potts model}
 \label{alg:robust tr}
\end{algorithm}

\subsection{Implementation details} \label{sec:impl details}
In all our experiments we used DeeplabV3+ \cite{deeplabv3plus} with MobileNetV2 \cite{sandler2018mobilenetv2} as a backbone model.

\textbf{Pretraining:} 
We use the standard ImageNet \cite{imagenet_cvpr09} pretraining of the backbone models. In addition, before the optimization via Grid-GD \eqref{eq:crf loss} and Grid-TR (\ref{eq:tr:s1}-\ref{eq:tr:grads2}) starts, the DeeplabV3+ models are pretrained by the PCE loss~\eqref{eq:pce loss}. 

\textbf{Meta-parameters:} 
We train $60$ epochs. We tuned the learning rates for all methods on the val set. Other meta-parameters for competitive methods were set as in the corresponding papers/code. The learning rate is polynomial with power $0.9$, momentum is $0.9$, batch size is $12$.

\textbf{Grid-TR \textsc{Stage A} \eqref{eq:tr:s1}:}
The low-level solver\footnote{GCOv3.0: \url{https://vision.cs.uwaterloo.ca/code/}} of the grid CRF is 
the $\alpha$-expansion \cite{boykov2001fast, Kolmogorov2004, Boykov2004} with 8-grid neighbourhood system. 
The max number of $\alpha$-expansion iterations is $5$ achieving convergence in most cases. We restrict the set of labels to those present in the image. We amortize the \textsc{Stage A} compute time by integrating it with data loading. The training is 1.3 times slower than Dense-GD.

\begin{figure*}
\begin{center}
    \newcommand{\wx}[1]{ \parbox{0.145\linewidth}{\centering #1} }
    \wx{image} \wx{ground truth}  \wx{PCE-GD \eqref{eq:pce loss}} \wx{Grid-GD} \wx{Dense-GD } \wx{Grid-TR (\ref{eq:tr:s1},\ref{eq:tr:grads2})} \\
    \includegraphics
    [width=0.96\linewidth,clip,trim=0 0 0 0]
    {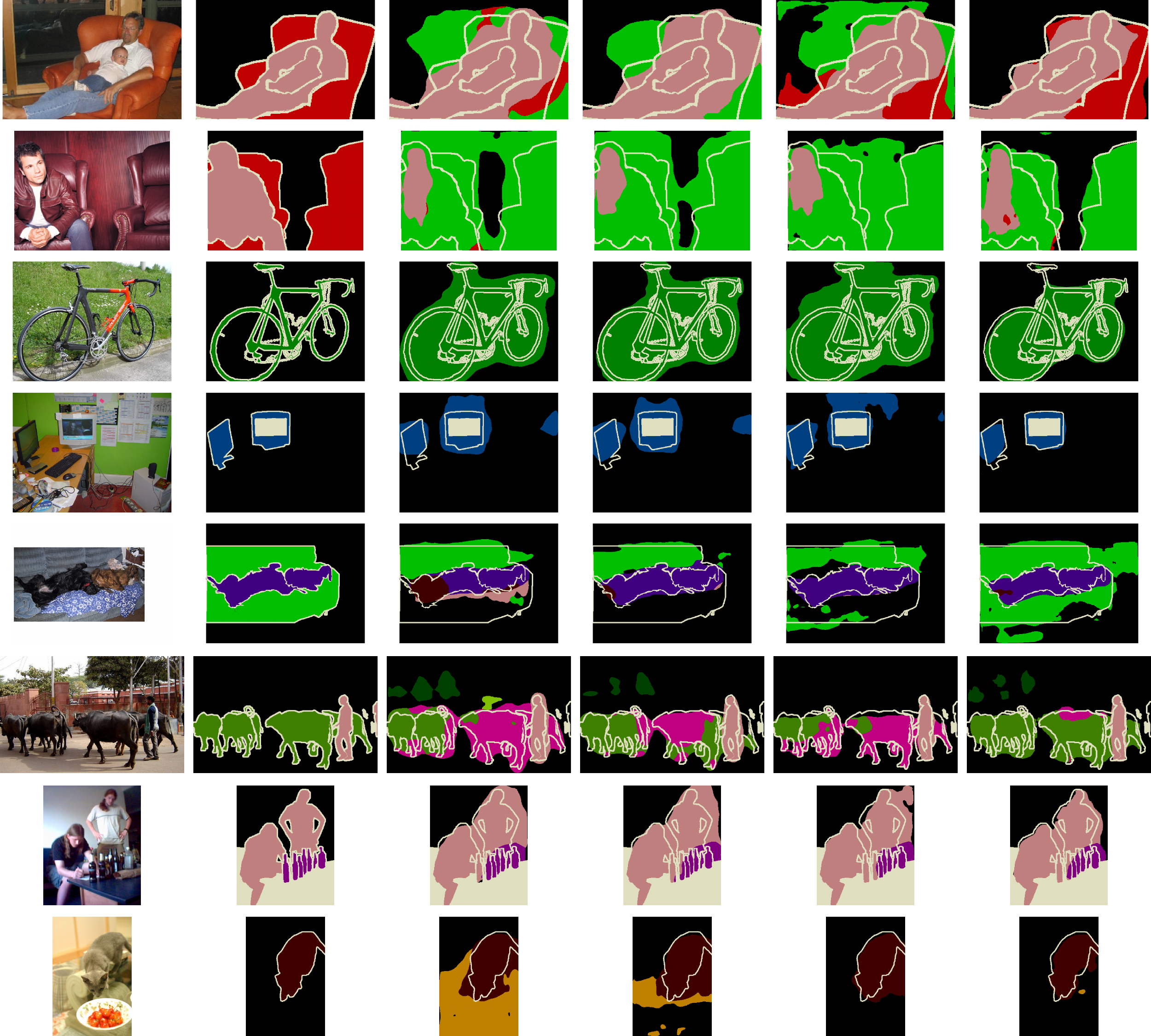}
\end{center}
    \caption{Examples of the full-scribble training results, see Tab.~\ref{tab:result} and Figure \ref{fig:long}. Note the better edge alignment of our Grid-TR.}
    \label{fig:examples}
\end{figure*}

\textbf{Grid-TR \textsc{Stage B} \eqref{eq:tr:grads2}:}
To amortize the time consumed by the graph cuts, we perform $M=5$ epochs of neural network weights updates \eqref{eq:tr:s2} for each update of the segmentation variables \eqref{eq:tr:grads2}. We use a global learning rate schedule spanning throughout iterations.
See Alg.\,\ref{alg:robust tr}.

\subsection{Segmentation quality}

\begin{table}[]
    \centering
    \small
    \begin{tabular}{l|rrrrr}
    \textbf{scribble length} & 0 & 0.3 & 0.5 & 0.8 & 1 \\
    \hline
    \hline
    full supervision & \multicolumn{5}{c}{0.70} \\
    \hline
    \textbf{PCE-GD} & 0.50 & 0.57 & 0.59 & 0.61 & 0.61 \\
    \textbf{Dense-GD} & 0.55 & 0.61 & 0.62 & 0.63 & 0.64 \\
    \textbf{Grid-GD} & 0.54 & 0.60 & 0.62 & 0.64 & 0.64 \\
    \textbf{Grid-TR} (our) & \textbf{0.57} & \textbf{0.63} & \textbf{0.64} & \textbf{0.66} & \textbf{0.67} \\
    \end{tabular}
    \caption{Results for ScribbleSup, see description in Figure \ref{fig:long}.}
    \label{tab:result}
\end{table}

\begin{figure}
\begin{center}
   \rotatebox{90}{\parbox{45mm}{\centering mean intersection\\[-0.5ex]over union (mIoU)}} \includegraphics[width=0.9\linewidth]{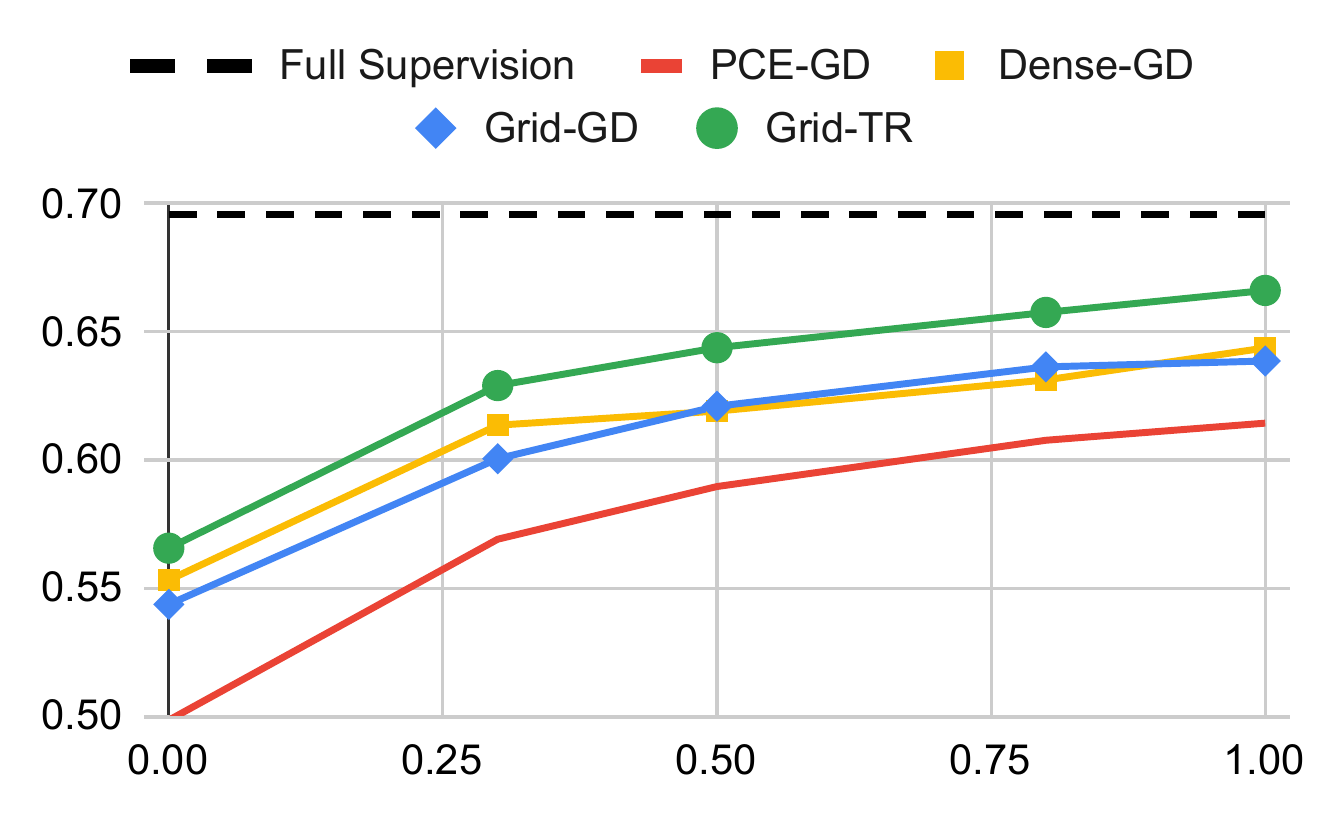}\\
   ~~~~~~~~~~~~scribble length ratio\\
\end{center}
   \caption{Segmentation performance on the val set of ScribbleSup \cite{scribblesup,pascal-voc-2012} using DeeplabV3+ \cite{deeplabv3plus} with MobileNetV2 \cite{sandler2018mobilenetv2} backbone. The supervision level varies horizontally, with $1$ corresponding to the full scribbles.
   Our ``Grid-TR'' outperforms other competitors for all scribble lengths and provides a new state-of-the-art.
   }
\label{fig:long}
\label{fig:onecol}
\end{figure}

\begin{figure}
\begin{center}
    \rotatebox{90}{\parbox{45mm}{\centering mean intersection\\[-0.5ex]over union (mIoU)}}%
    \includegraphics[width=0.9\linewidth]{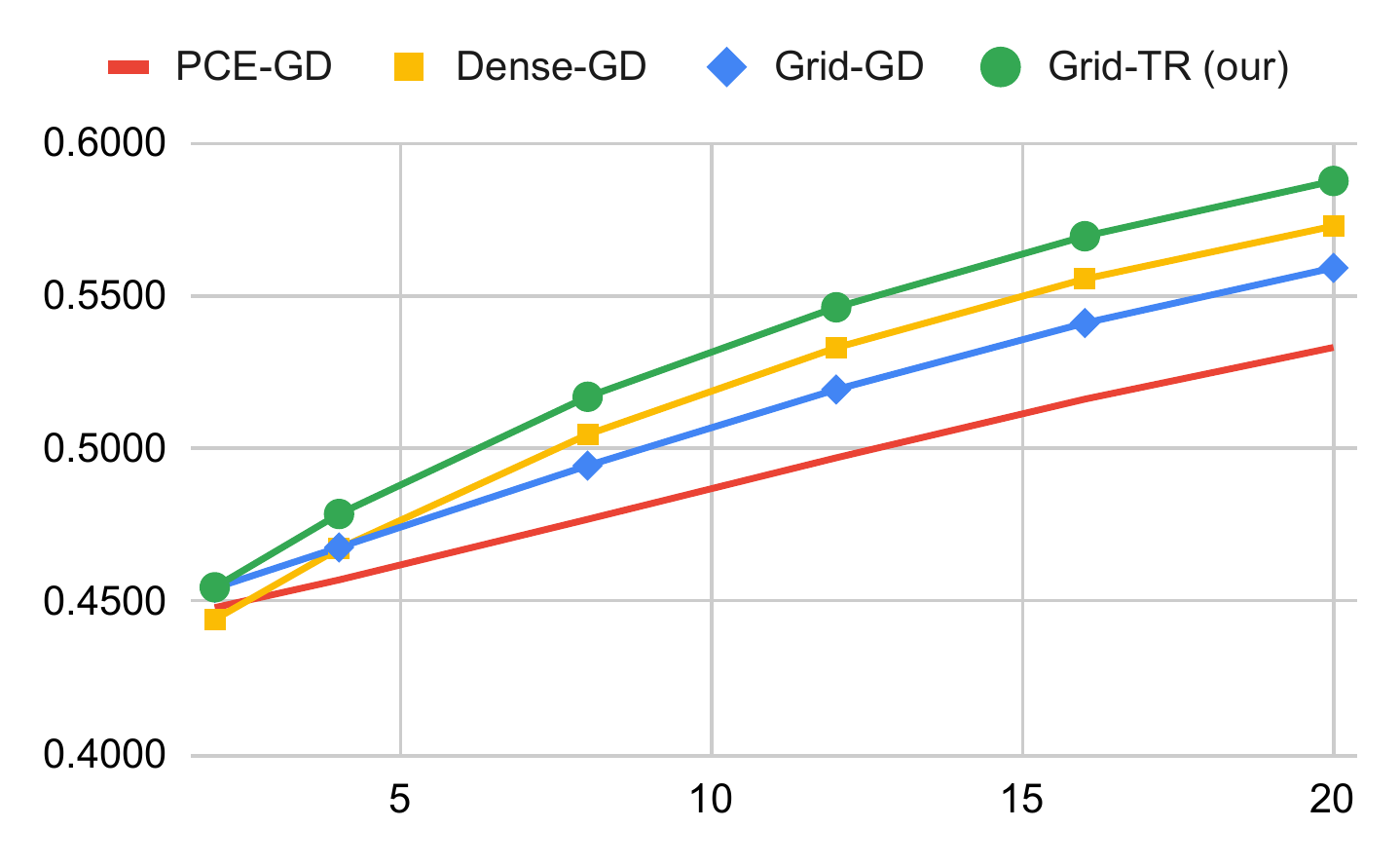}\\
   ~~~~~~distance to the boundary (trimap width), px 
\end{center}
    \caption{The quality of segment boundary alignment. The networks were trained on the full-length scribbles.}
    \label{fig:trimaps}
\end{figure}

The quantitative results of the weakly supervised training for semantic segmentation are presented in Figure \ref{fig:long} and Tab.~\ref{tab:result}. The results are presented with different levels of supervision varying from the clicks (denoted as length $0$) to the full-length scribbles (denoted as length $1$). Decreasing supervision results in degraded performance for all methods. We are interested to compare how different approaches perform at different levels of supervision. Our Grid-TR outperforms all the competitors at each level of supervision.

The examples of images and results shown in Fig.~\ref{fig:examples} demonstrate the advantages of our method, particularly \wrt edge alignment. Quantitatively, we evaluate the accuracy of semantic boundaries using standard trimaps \cite{kohli2009robust, koltun:NIPS11, deeplab, marin2019efficient}. A trimap corresponds to a narrow band around the ground truth segment boundaries of varying width. An accuracy measure, \eg mIoU, is computed for pixels within each band. The results are shown in Fig.~\ref{fig:trimaps} where our approach demonstrates superior performance.

\section{Discussion}
\subsection{On parameter $\lambda$ in \eqref{eq:tr:s1} }    %
\label{sec:discussion}

\begin{figure}
    \centering
    \rotatebox{90}{\parbox{45mm}{\centering mean intersection\\[-0.5ex]over union (mIoU)}}%
    \includegraphics[width=0.9\linewidth]{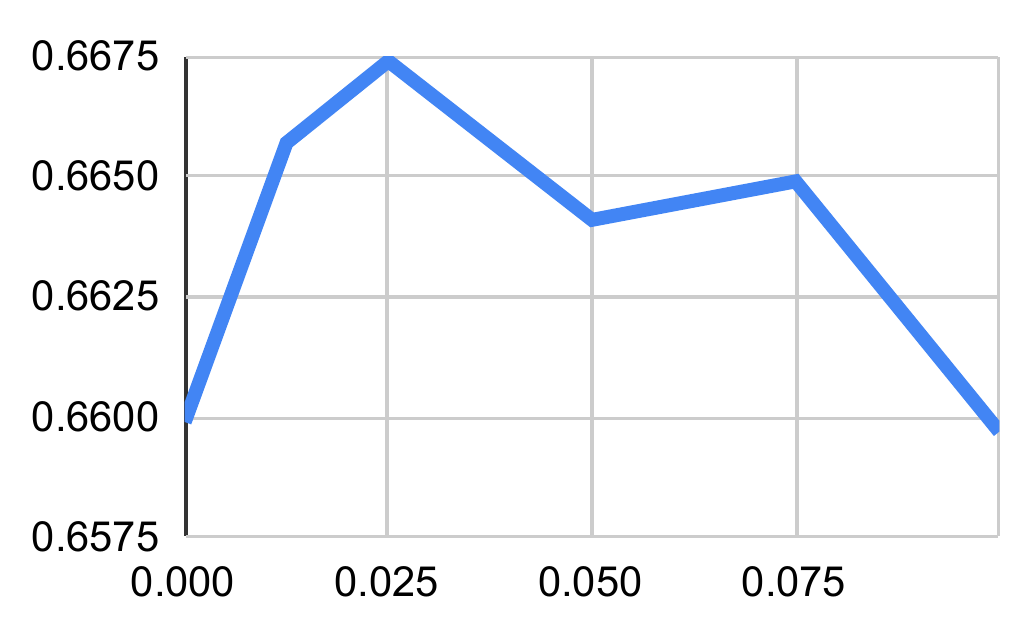} \\[-2ex]
    \hspace{1cm} value of $\lambda$ - Lagrange multiplier in \eqref{eq:tr:s1} \\[1ex]
    \caption{Empirical evaluation of Lagrange multiplier $\lambda$ for the Trust Region term in \eqref{eq:tr:s1}: the plot shows how
    mobile-net training quality depends on $\lambda$. The context is the weakly-supervised semantic segmentation in Sec.4 with 
    regularization loss $E$ (using 8-grid Potts and full scribbles) based on our Trust Region chain rule with robust metric 
    $\dB$ in \eqref{eq:tr:grads2}. For $\lambda=0$ equation \eqref{eq:tr:s1} generates fixed low-level segmentation proposals completely independent 
    of the network. Then, the network overfits to mistakes in such proposals due to biases/weaknesses of the regularizer. 
    As $\lambda\rightarrow \infty$, trust region becomes too small and our approach loses its advantages 
    due to better (\eg higher-order) approximation $\tilde{E}$ in \eqref{eq:tr:s1}. %
    Conceptually speaking, it should get closer to the results of gradient descent, which uses basic first-order approximations.} 
    \label{fig:my_label}
\end{figure}

As discussed below equations \eqref{eq:tr:s1} - \eqref{eq:tr:grads2} in the paper, even for exact (global) solvers 
using $\tilde{E}=E$ in \eqref{eq:tr:s1}, the choice of $\lambda=0$ could be sub-optimal, %
as demonstrated empirically here in Figure \ref{fig:my_label}. As argued in the paper,
while $\lambda=0$ with an exact solver may seem like a good approach to training $\min_\theta E(\ff(\theta))$
suggesting globally optimal loss, empirically this leads to overfitting to mistakes or biases of the
regularizer (e.g. the Potts model). One argument for $\lambda>0$ discussed in the paper is that this 
corresponds to the constrained optimization of \eqref{eq:Ef} %
over the network manifold in $\R^n$.
Such formulation of the training could be preferred as constraining to neural networks %
can be seen as incorporation of the ``deep priors", \eg \cite{ulyanov2018deep}. One can also argue that
local minima of $E$ inside the manifold of the network output in $\R^n$ may be preferable to the global optimum of $E$
due to limitations of the basic (but solvable) regularizers.

Empirically, $\lambda=0$ in \eqref{eq:tr:s1} leads to a fixed set of proposals generated in a single run of stage A
completely independent of the network. In contrast, $\lambda>0$ leads to multiple distinct iterations of stage A
where the network is in the feedback loop. Vice versa, instead of fixed proposals, for $\lambda>0$ 
the network is exposed  to a substantially larger set of solutions in stage B reduces overfitting.

Moreover, the objective in \eqref{eq:tr:s1} %
can be motivated on its own merits independently of the objective in \eqref{eq:Ef}. %
It can be seen as a low-level segmentation objective that integrates
class likelihoods produced by the neural network, replacing the basic likelihoods using low-level features, 
\eg colors, as discussed in Sec.1.1. %
Iterations A/B can be seen as joint segmentation and model estimation, as typical for
well-known low-level segmentation methods like Zhu-Yuille \cite{Zhu:96}, Chan-Vese \cite{Chan-Vese-01b}, or GrabCut \cite{GrabCuts:SIGGRAPH04}. The main difference is that our stages A/B use ``deep'' models. In contrast to
standard methods \cite{Zhu:96,Chan-Vese-01b,GrabCuts:SIGGRAPH04} estimating model parameters for 
some standard class of probability distributions (\eg GMM) over fixed low-level features like colors, 
we estimate deep models with millions of parameters that can be interpreted as learning high-level (semantic) features.

\subsection{On discrete losses and decisions/activations} \label{sec:discussion2}

Our approach can train networks using discrete decisions/activations and
losses defined over discrete domains. For example, \eqref{eq:tr:s1}-\eqref{eq:tr:grads2}  %
do not require that $E$ is differentiable. In particular, \eqref{eq:tr:s1} %
can be optimized over ``hard" segmentations $\fo\in\{0,1\}^{N\times K}\subset\Delta_K^N$ even if the network 
produces soft segmentations $\ff(\theta)\in\Delta_K^N$, as long as $\dA$ in \eqref{eq:tr:s1} can measure a distance 
between discrete and continuous solutions, \eg $\KL(s,\ff)$ for one-hot and soft distributions.
It is also possible to train the models with discrete decision functions $\mathbf D(l)$ such that 
$\ff(\theta)= {\mathbf D}(l(\theta))$ where $l$ are logits. Then, all arguments in \eqref{eq:tr:s1} are discrete.
Optimization in \eqref{eq:tr:s2} can be formulated over real-valued logits using $\dB$ measuring a distance to subset 
$\{l\,|\,{\mathbf D}(l)=s_{t+1}\}\subset\R^{N\times K}$.

\section*{Acknowledgements}

We thank Yaoliang Yu for the insightful discussion on related proximal methods and pointing out related literature. We also thank Vladimir Kolmogorov for 
suggesting prior studies of the tightness of the Potts model relaxations.

\input{main_arxiv2.bbl}
\end{document}

%% file: images/graphical_model.tex
\usetikzlibrary{bayesnet}
\usetikzlibrary{arrows}
\usetikzlibrary{backgrounds}
\usetikzlibrary{calc}

% \begin{document}
 
\tikz{
% nodes
 \node[obs] (i) {$I$};%
 \node[yshift=8mm] (i-d) at (i) {\parbox{2cm}{\centering\textbf{\textit{\small observed\\[-1ex]image}}}};
 
 \node[latent,right=of i,fill] (z) {$Z$}; %
 \node[yshift=8mm] (z-d) at (z) {\parbox{2cm}{\centering\textbf{\textit{\small hidden\\[-1ex]true\\[-1ex]labeling}}}};
 
 \node[obs,right=of z] (y) {$Y$}; %
 \node[yshift=8mm] (y-d) at (y) {\parbox{2cm}{\centering\textbf{\textit{\small observed\\[-1ex]noisy\\[-1ex]labeling}}}};
 
 \node (middle) at ($(z)!0.5!(y)$) {};
 \node[yshift=-8mm] (middle2) at ($(z)!0.5!(y)$) {};
 \node[yshift=-9mm] (process) at (z) {\small switch label with probability $\varepsilon$};
 
 % plate
 \plate [inner sep=.25cm] {plate1} {(i-d)(z-d)(y-d)(y)(middle2)} {}; %
% edges
 \edge {i} {z};
 \edge {z} {y}; 
 \draw [-stealth,dotted] (middle2) -- (middle); 
}

% \end{document}

%% file: images/robust_plot.tex
\definecolor{Chartreuse}{RGB}{190, 190, 0}
\begin{tikzpicture}
  \draw[->] (-2, 0) -- (2, 0) node[right] {$x_i$};
  \draw[->] (0, -0.2) -- (0, 2) ;
  \draw (1, 0.1) -- (1, -0.1) node[below] {\scriptsize $1$};
  \draw[scale=0.5, domain=-4:4, thick, smooth, variable=\x, Chartreuse] plot ({\x}, {ln(1+exp(-\x))});
  \draw[scale=0.5, domain=-4:4, thick, smooth, variable=\x, blue] plot ({\x}, {-ln(0.1 + 0.8 / (1+exp(-\x)))});
  
  \node at (-0.8, 1.7) {\color{Chartreuse}\scriptsize (a): $-\log q^1_i$ };
  \node at (-1.8, 0.5) {\color{blue}\scriptsize (b): $-\log (a + b\,q^1_i)$ };
  \node at (1.2, 1) {\scriptsize $q^1_i = \frac1{1+\exp(-x_i)}$ };
\end{tikzpicture}